\pdfoutput=1

\documentclass[11pt]{article}

\usepackage{acl}

\usepackage{times}
\usepackage{latexsym}
\usepackage{booktabs}
\usepackage{graphicx}
\usepackage{multirow}
\usepackage{pgfplots}
\usepackage{pgf-pie}
\usepackage{todonotes}
\usepackage{lscape}
\usepackage{subfig}
\usepackage{filecontents}
\usepackage{graphicx}
\usepackage{xcolor}
\usepackage{wrapfig}
\usepackage{listings}
\usepackage{xcolor}
\usepackage{graphicx}
\usepackage{subcaption}

\usepackage[T1]{fontenc}

\usepackage[utf8]{inputenc}

\usepackage{microtype}

\usepackage{inconsolata}

\usepackage{geometry}
\usepackage{array}
\usepackage{longtable}
\usepackage{makecell}
\usepackage[justification=centering]{caption}
\definecolor{codegray}{gray}{0.9}
\definecolor{mygreen}{RGB}{126, 172, 85}
\definecolor{myred}{RGB}{147, 29, 20}

\newcommand{\code}[1]{\colorbox{codegray}{\texttt{#1}}}

\lstdefinestyle{json}{
    basicstyle=\ttfamily\small,
    commentstyle=\color{gray},
    stringstyle=\color{blue},
    keywordstyle=\color{red}
}

\title{\texttt{PARADISE}: Evaluating Implicit Planning Skills of \\ Language Models with Procedural Warnings and Tips Dataset}

\author{Arda Uzunoğlu\textsuperscript{$\ddagger$},~
Abdulfattah Safa\textsuperscript{$\dagger$},~
Gözde Gül Şahin\textsuperscript{$\dagger$}~
\\[.3em]
\textsuperscript{$\ddagger$}Computer Science Department, Johns Hopkins University, Maryland, USA \\
\textsuperscript{$\dagger$}Computer Engineering Department, Koç University, Istanbul, Türkiye\\
\textsuperscript{$\dagger$}{KUIS AI, Koç University, Istanbul, Türkiye}\\
\textsuperscript{$\dagger$}{\url{https://gglab-ku.github.io/}}\\
}

\begin{document}
\maketitle
\begin{abstract}
Recently, there has been growing interest within the community regarding whether large language models are capable of planning or executing plans. However, most prior studies use LLMs to generate high-level plans for simplified scenarios lacking linguistic complexity and domain diversity, limiting analysis of their planning abilities. 
These setups constrain evaluation methods (e.g., predefined action space), architectural choices (e.g., only generative models), and overlook the linguistic nuances essential for realistic analysis. 
To tackle this, we present \textsc{PARADISE}, an abductive reasoning task using Q\&A format on practical procedural text sourced from wikiHow. It involves warning and tip inference tasks directly associated with goals, excluding intermediary steps, with the aim of testing the ability of the models to infer implicit knowledge of \textit{the plan} solely from the given goal. Our experiments, utilizing fine-tuned language models and zero-shot prompting, reveal the effectiveness of task-specific small models over large language models in most scenarios. Despite advancements, all models fall short of human performance. Notably, our analysis uncovers intriguing insights, such as variations in model behavior with dropped keywords, struggles of BERT-family and GPT-4 with physical and abstract goals, and the proposed tasks offering valuable prior knowledge for other unseen procedural tasks. The \textsc{PARADISE} dataset and associated resources are publicly available for further research exploration with \url{https://github.com/GGLAB-KU/paradise}.

\end{abstract}

\section{Introduction}
\label{sec:introduction}
\begin{figure}[h]
    \centering
    \includegraphics[width=0.5\textwidth]{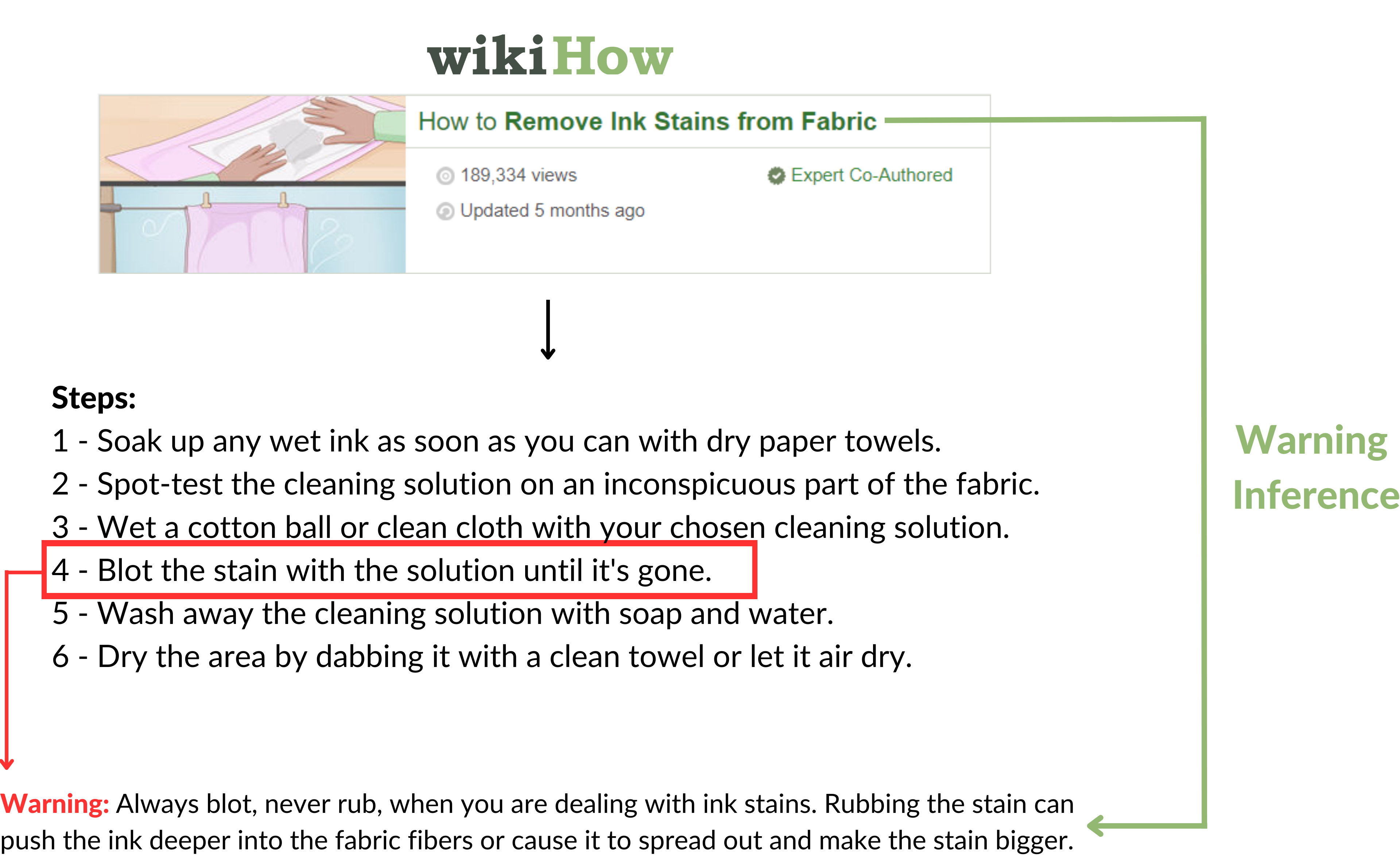}
    \captionsetup{justification=raggedright,singlelinecheck=false}
    \caption{A procedural tutorial on ``Removing Ink Stains from Fabric''. Here, one can damage the fabric if they ignore the warning ``Always blot, never rub, when dealing with ink stains''.}
    \label{figure:warning_inference}
\end{figure}

Recent breakthroughs in emergent (or lack of) abilities of large language models~(LLMs) have given rise to empirical studies that employ language models as planners~\cite{HuangAPM22,zhao2023large,SongSWCW023} (i.e., agentic models) and analysis studies that investigate their planning capabilities~\cite{criticalInvest23,understandingPlanning,planningAboutChange}. Majority of these studies use toy simulation environments like ALFRED~\cite{ShridharTGBHMZF20}, BlocksWorld, and VirtualHome~\cite{PuigRBLWF018} which have little lexical and domain variance and limited number of actions (e.g., verbs). Additionally, the planning task is mostly formulated as a generation problem which can only be evaluated on the closed problem domain; and models with decoder-based architectures. Hence evaluating \textit{open-domain} planning abilities for a wide range of models still remain as a challenge.        

Planning requires a combination of a wide range of complex reasoning abilities. One line of research focuses on distinct set of reasoning abilities (e.g., commonsense~\citep{huang-etal-2019-cosmos}, arithmetic~\cite{gsm8k}, logical~\cite{han2022folio}, temporal~\cite{tram23} etc...) of language models on more realistic, open-domain text. However, such open-domain text mostly does not contain \textit{a goal} or \textit{a plan}, hence lack the complex linguistic phenomenon (e.g., implicit relations, complex temporal, and co-reference links, large event-cause chains etc...) that is common in procedural text. On the other hand, existing studies that utilize plans as a test bed for reasoning are mostly limited to extracting direct and explicit relations~\cite{dalvi-etal-2019-everything,zhang-etal-2020-reasoning}, i.e., more simplistic reasoning compared to implicit reasoning.  
Here, we formulate the implicit planning skills as abductive reasoning skills over the procedural plans with missing information. We hypothesize that a model with planning abilities would be able to \textit{infer the warnings and tips} about the plan \textit{without seeing the explicit instructions} (see Fig~\ref{figure:warning_inference}). 
Although abductive reasoning has been studied in the past, they either employ additional source of information \citep{huang-etal-2019-cosmos, bhagavatula2020abductive} or focus on consecutive elements \citep{zellers-etal-2018-swag, zellers-etal-2019-hellaswag, tandon-etal-2019-wiqa}, both of which diminish the notion of implicity. 

In this work, we present \textsc{PARADISE}, an extensive, expert-curated dataset for warning and tip inference tasks covering a wide range of domain derived from wikiHow\footnote{\url{https://www.wikihow.com}}. Unlike previous works, our tasks focus on the implicit relationship between goals and warnings/tips, bypassing intermediate steps (instructions). This requires a model to possess implicit knowledge of intermediate steps (i.e., the plan) solely based on the provided goal in the absence of explicit instructions. Furthermore, we use a question answering formulation, which allows for easier evaluation with standard metrics, and testing of broader-range of model architectures including encoder-based ones. 
We establish robust baselines by fine-tuning pretrained language models like DeBERTa \citep{deberta} and zero-shot prompting with a varied set of large language models such as Mistral-7B \citep{jiang2023mistral} and GPT-4 \citep{openai2023gpt4}. Our extensive experiments address a broad range of research questions, delving into the relationship between memorization and performance~(\S \ref{subsection:keyword_manipulation}), the differences in failures between PLMs and LLMs ~(\S \ref{subsection:plm_llm_comparison}), and the knowledge transfer capacity of the proposed tasks to unseen tasks ~(\S \ref{subsection:transfer_learning}). We observe that fine-tuning small models tailored to specific tasks proves more effective than zero-shot prompting across all LLMs, including GPT-4. However, it's noteworthy that all models, despite these efforts, still lag behind human performance. Our exhaustive analysis also provide interesting insights such as large models getting less affected from dropping matched keywords; BERT-family struggling more with physical goals, while GPT-4 struggling with abstract, digital and social objectives; and proposed tasks providing beneficial prior knowledge to unseen procedural tasks. We release all the resources at \url{https://github.com/GGLAB-KU/paradise}. 

\section{PARADISE}
\label{sec:dataset}
Initially, we augment the wikiHow corpus~\citep{zhang-etal-2020-reasoning} by integrating it with a recent compilation\footnote{Scrape date: November, 2022} of 21K tutorials. The extended corpus maintains the JSON format, except for the segregation of warnings and tips, an example of which can be seen in Appendix \ref{sec:example_json_file}. Each wikiHow tutorial comprises procedural steps to achieve its objective, with some tutorials featuring step-specific or general warnings and tips. We automatically generate downstream task data incorporating these warnings and tips, as elaborated in Sec.~\ref{subsection:candidate_sampling}. The process also involves expert human annotation, detailed in Sec.~\ref{subsection:test_set_construction}.

\subsection{Task Formulation}
\label{subsection:task_formulation}

We define warning and tip inference tasks as multiple-choice question answering tasks, in which a system needs to choose the correct warning or tip for a given goal among candidates. In this context, goals are questions, while warnings and tips are the choices. An example for both tasks can be seen in Fig.~\ref{fig:examples}.

\begin{figure}
  \centering
  \resizebox{0.7\textwidth}{!}{%
    \begin{minipage}{\textwidth}
      \begin{enumerate}
        \item Goal: Sit up Straight at a Computer
        \begin{enumerate}
          \item Remember that people can see some of \\ your surroundings you while you chat. Be \\ mindful of what is in the camera's field of view.
          \item \textbf{Do not remain in any one position in front of \\ a computer for too long.}
          \item Avoid moving around in this pose. Any \\ movements you make within the pose should \\ be deliberate and slow.
          \item Keep an appropriate distance between your \\ eyes and computer screen.
        \end{enumerate}
        \item Goal: Avoid Oil Splatter when Frying
        \begin{enumerate}
          \item Remember to have lots of sides \\ apart from just the barbecued food.
          \item Wear clear, plastic gloves if you are going \\ to use your hands to mix the meat.
          \item Never use extra virgin olive oil to stir-fry. \\ It has a low smoking point.
          \item \textbf{Wear long sleeves when you plan on \\ frying food.}
        \end{enumerate}
      \end{enumerate}
    \end{minipage}%
  }
  \captionsetup{justification=raggedright,singlelinecheck=false}
  \caption{Example questions for warning (1) and tip (2) inference tasks. Correct choices are \textbf{bold}.}
  \label{fig:examples}
\end{figure}

\begin{table*}[!htp]
            \begin{center}
            \scalebox{0.78}{
            \begin{tabular}{lccccccccc|ccc}
            \toprule
             & \multicolumn{9}{c}{\textbf{Category Distribution}} & \multicolumn{3}{c}{\textbf{Size}} \\
             \midrule
             & \textbf{Other} &\textbf{C\&E} & \textbf{HE} & \textbf{F\&E} & \textbf{H\&C} & \textbf{H\&G} & \textbf{E\&C} & \textbf{F\&B} & \textbf{PC\&S} & \textbf{Train} & \textbf{Dev} & \textbf{Test}\\
            \midrule
            WikiHow Corpus & 27.25\% & 14.51\% & 10.18\% & 10.09\% & 9.27\% & 8.89\% & 7.35\% & 6.63\% & 5.83\% & \multicolumn{3}{c}{133K} \\
            Warning Inference & 34.75\% & 7.09\% & 14.13\% & 7.26\% & 4.12\% & 14.02\% & 5.19\% & 4.92\% & 8.52\% & 33K & 5K & 500 \\
            Tip Inference & 30.13\% & 7.45\% & 11.33\% & 10.61\% & 5.44\% & 12.11\% & 7.46\% & 5.29\% & 10.18\% & 71K & 5K & 500\\
            \bottomrule
            \end{tabular}
            }
            \captionsetup{justification=raggedright,singlelinecheck=false}
            \caption{Category distribution and size of \textsc{PARADISE}. C\&E - Computer and Electronics, HE - Health, F\&E - Food and Entertaining, H\&C - Hobbies and Crafts, H\&G - Home and Garden, E\&C - Education and Communications, F\&B - Finance and Business, PC\&S - Personal Care and Style.}
            \label{table:dataset_stats}
            \end{center}
\end{table*}

\subsection{Candidate Sampling}
\label{subsection:candidate_sampling}

Acquiring the goals and positive candidates is straightforward, involving iterative selection from each tutorial in our corpus. For negative candidate sampling, we modify the approach outlined by \citet{zhang-etal-2020-reasoning}. In contrast to their reliance solely on verbs, we note that verbs prove inadequate in capturing meaning because warnings and tips, on average, are much longer than individual steps (\textasciitilde 40 versus \textasciitilde 11 tokens). This leads to the generation of low-quality negative candidates. To address this limitation, we enhance our negative candidate sampling strategy by incorporating embeddings of noun tokens. 

We begin by encoding each warning and tip using BERT~\citep{devlin-etal-2019-bert}. We calculate the average of verb and noun tokens, identified with spaCy~\citep{Honnibal_spaCy_Industrial-strength_Natural_2020}. Subsequently, we employ FAISS~\citep{johnson2019billion} to conduct a semantic similarity search, identifying the top three warnings and tips with the highest cosine similarity score relative to the positive candidate.

Following \citet{zhang-etal-2020-reasoning}, we randomly reassign one of the negative candidates as positive and correct the labels and goals accordingly with a probability of 0.15 to avoid sampling bias and filter the examples as described in Appendix \ref{sec:filtering_examples}.

\subsection{Test Set Construction}
\label{subsection:test_set_construction}

As our datasets are automatically generated, they may include undesired elements like multiple plausible candidates for a given goal. For instance, consider the goal ``Deal with Your Step Mother'', which has a positive candidate of ``Stay connected with relatives such as grandparents and close friends for extra support'' and a negative candidate of ``Recruit help from friends and family''. Although the negative candidate is chosen due to its high semantic similarity with the positive candidate, it is also a reasonable choice for the given goal, introducing noise into the dataset. To mitigate such issues, we employ expert annotation to validate the test splits.

The expert annotation process consists of three stages. First, experts verify that each example contains no more than one plausible candidate. Second, they meticulously examine examples to ensure that the positive candidate is genuinely relevant to the content of the associated wikiHow tutorial. Finally, experts assess the appropriateness of examples for gauging reasoning skills, excluding instances that demand expert-level knowledge or domain-specific high-level information. This annotation process yields approximately 80\% of annotations as valid examples. Consequently, the test splits for each task are expanded with such valid examples from the pool of automatically generated examples until reaching the predetermined size of 500 examples.

Apart from expert annotation, we leverage the dataset cartography tool~\citep{swayamdipta-etal-2020-dataset} to uphold the high quality of our data, probe our datasets, and gain a deeper understanding of their features. Further details can be found in Appendix~\ref{sec:dataset_analysis_with_cartography}.

\subsection{Dataset Statistics}
\label{subsection:dataset_statistics}

The statistics of the corpus and final datasets are given in Table~\ref{table:dataset_stats}. We specify the validation and test split sizes as 5K and 500, respectively, with the remaining data serving as the training set. Tip inference dataset is nearly twice the size of the warning inference dataset, but the average token counts per goal and candidate are comparable~(\textasciitilde 7 for goal, \textasciitilde 40 for candidate). We employ a nearly uniform sampling approach across various categories to ensure a high level of domain diversity.

\section{Experimental Setup}
\label{sec:experimentsal_setup}


To evaluate language models in our tasks, we establish two setups: 1) finetuning setup for pretrained encoder LMs such as BERT~\citep{devlin-etal-2019-bert} and 2) zero-shot setup for large language models such as GPT-4~\citep{openai2023gpt4}. 

\subsection{Finetuning Setup}
\label{subsection:plms}
We fine-tune a set of models from the BERT family: DistilBERT \citep{Sanh2019DistilBERTAD}, BERT \citep{devlin-etal-2019-bert}, RoBERTa \citep{Liu2019RoBERTaAR}, DeBERTa \citep{deberta}, which show strong performance in procedural and reasoning tasks~\citep{zhou-etal-2022-show, tandon-etal-2019-wiqa, zellers-etal-2019-hellaswag, zhang-etal-2020-reasoning}. For fine-tuning, we concatenate each candidate (warning or tip) with the question (goal) using a \code{[CLS]} token, i.e., the model receives \code{[CLS]} \texttt{question} \code{[SEP]} \texttt{candidate} as input.  Subsequently, we apply an additional projection layer, followed by a softmax function that receives the representation of the \code{[CLS]} token for each \texttt{candidate}. The \texttt{candidate} with the highest probability is selected as the answer. The models are optimized through cross-entropy loss. Further implementation details can be found in Appendix \ref{sec:implementation_details}.


\subsection{Zero-Shot Setup}
\label{subsection:llms}

We identify five popular and capable\footnote{These LLMs are chosen from the models that rank high in the \hyperlink{https://huggingface.co/spaces/lmsys/chatbot-arena-leaderboard}{Hugging Face Chatbot Leaderboard}.} large language models that are diverse in architecture, scale, avalability, and performance, namely as GPT-4\footnote{Model variant: \texttt{gpt-4-1106-preview}}~\citep{openai2023gpt4}, PALM-2 \footnote{Model variant: \texttt{text-bison}}~\citep{anil2023palm}, LLaMA-2 70B \cite{touvron2023llama}, Mistral 7B\footnote{Model variant: \texttt{\href{https://huggingface.co/mistralai/Mistral-7B-Instruct-v0.1}{Mistral-7B-Instruct-v0.1}}}~\citep{jiang2023mistral}, and Vicuna 33B~\citep{vicuna2023}.

We first perform preliminary experiments with default prompts on a small subset of the validation set. We, then, iteratively refine the prompts to fit the specific model's template. For instance Vicuna expects a certain template with \code{\#\#\#Human} and \code{\#\#\#Assistance} roles specified in the text. We use the respective model APIs, where available. Preliminary tests on the subsets were conducted for each model to identify optimal \texttt{temperature} and \texttt{top\_p} parameters. The best-performing configurations were then applied to the entire datasets for a thorough evaluation. Further details on the prompt templates and parameters are given in Appendix \ref{sec:prompting_llms}.

\section{Experiments and Results}
\label{sec:experiments_and_results}
\begin{table}
    \begin{center}
    \scalebox{0.9}{
        \begin{tabular}{lccc}
            \hline
            {\textbf{Model}} & &
                {\textbf{Warning}} &
                {\textbf{Tip}} \\
                \toprule
                Random & & 25.0 & 25.0 \\
                Majority & & 26.0 & 26.0 \\
                \midrule
                \multicolumn{4}{c}{\textbf{PLMs}} \\
                \midrule
                DistilBERT &  & 22.44 \textpm 3.88 & 21.48 \textpm 4.84 \\
                BERT & & \textbf{25.52 \textpm 4.56} & \textbf{26.88 \textpm 5.02} \\
                RoBERTa & & 20.88 \textpm 4.80 & 20.36 \textpm 4.20 \\
                DeBERTa & & 23.40 \textpm 6.76 & 22.60 \textpm 8.61 \\
                \midrule
                \multicolumn{4}{c}{\textbf{Fine-Tuned PLMs}} \\
                \midrule
                DistilBERT & & 82.16 \textpm 0.79 & 87.48 \textpm 0.65 \\
                BERT & & 83.92 \textpm 0.68 & 89.80 \textpm 0.91 \\
                RoBERTa & & 87.92 \textpm 0.60 & 91.00 \textpm 0.34 \\
                DeBERTa & & \textbf{90.68 \textpm 0.41} & \textbf{93.68 \textpm 0.48} \\
                \midrule
                \multicolumn{4}{c}{\textbf{Open-Source LLMs}} \\
                \midrule
                Mistral 7B & & \textbf{71.8} & \textbf{72.4} \\
                Vicuna 33B & & 53.2 & 57.0 \\
                LLaMA-2 70B & & 65.2 & 64.5 \\
                \midrule
                \multicolumn{4}{c}{\textbf{ Proprietary LLMs}} \\
                \midrule
                PALM-2 & & 83.6 & 82.4 \\
                GPT-4 & & \textbf{86.2} & \textbf{88.8} \\
                \midrule
                \textbf{Human} & & \textbf{94.0} &\textbf{96.0} \\
                \bottomrule
        \end{tabular}}
    \captionsetup{justification=raggedright,singlelinecheck=false}
    \caption{Main accuracy results for fine-tuning and zero-shot setups.}
    \label{table:inference_tasks}
    \end{center}
\end{table}

\begin{figure*}[ht]
\begin{centering}
\resizebox{\textwidth}{!}{%
    \begin{tikzpicture}
        \begin{axis}[
        title={Warning Inference},
        xlabel={},
        ylabel={Accuracy},
        ybar=1.0pt,
        bar width=7pt,
        symbolic x coords={DistilBERT, BERT, RoBERTa, DeBERTa, Mistral 7B, Vicuna 33B, LLaMA-2 70B, PALM-2, GPT-4},
        grid=both,
        ymin=0,
        legend cell align=left,
        enlarge x limits=0.05,
        ymin=0,
        ymax=1,
        x label style={font=\footnotesize},
        y label style={font=\footnotesize},
        ticklabel style={font=\footnotesize},
        width=16cm,
        height=8cm,
        ]
        \addplot[black,fill=red,error bars/y dir=both,error bars/y explicit] coordinates {
          (DistilBERT, 0.8216) +- (0, 0.0079)
          (BERT, 0.8392) +- (0, 0.0068)
          (RoBERTa, 0.8792) +- (0, 0.0060)
          (DeBERTa, 0.9068) +- (0, 0.0041)
          (Mistral 7B, 0.718)
          (Vicuna 33B, 0.532)
          (LLaMA-2 70B, 0.652)
          (PALM-2, 0.836)
          (GPT-4, 0.862)
        };
        \addplot[black,fill=violet,error bars/y dir=both,error bars/y explicit] coordinates {
          (DistilBERT, 0.6176) +- (0, 0.0092)
          (BERT, 0.6492) +- (0, 0.0093)
          (RoBERTa, 0.7200) +- (0, 0.0099)
          (DeBERTa, 0.7660) +- (0, 0.0081)
          (Mistral 7B, 0.556)
          (Vicuna 33B, 0.431)
          (LLaMA-2 70B, 0.414)
          (PALM-2, 0.758)
          (GPT-4, 0.810)
        };
      \end{axis}
    \end{tikzpicture}
    \begin{tikzpicture}
      \begin{axis}[
        title={Tip Inference},
        xlabel={},
        ylabel={},
        set layers,
        ybar=1.0pt,
        bar width=7pt,
        symbolic x coords={DistilBERT, BERT, RoBERTa, DeBERTa, Mistral 7B, Vicuna 33B, LLaMA-2 70B, PALM-2, GPT-4},
        grid=both,
        ymin=0,
        legend cell align=left,
        enlarge x limits=0.05,
        ymin=0,
        ymax=1,
        x label style={font=\footnotesize},
        y label style={font=\footnotesize},
        ticklabel style={font=\footnotesize},
        legend image post style={scale=1.00},
        legend style={at={(1.05,0.1)},anchor=west,font=\scriptsize},
        width=16cm,
        height=8cm,
        ]
        \addplot[black,fill=red,error bars/y dir=both,error bars/y explicit] coordinates {
          (DistilBERT, 0.8748) +- (0, 0.0065)
          (BERT, 0.8980) +- (0, 0.0091)
          (RoBERTa, 0.9100) +- (0, 0.0034)
          (DeBERTa, 0.9368) +- (0, 0.0048)
          (Mistral 7B, 0.724)
          (Vicuna 33B, 0.57)
          (LLaMA-2 70B, 0.645)
          (PALM-2, 0.824)
          (GPT-4, 0.888)
        };
        \addlegendentry{Original}
        \addplot[black,fill=violet,error bars/y dir=both,error bars/y explicit] coordinates {
          (DistilBERT, 0.6624) +- (0, 0.0154)
          (BERT, 0.69) +- (0, 0.0108)
          (RoBERTa, 0.7436) +- (0, 0.0046)
          (DeBERTa, 0.8072) +- (0, 0.0144)
          (Mistral 7B, 0.592)
          (Vicuna 33B, 0.455)
          (LLaMA-2 70B, 0.555)
          (PALM-2, 0.778)
          (GPT-4, 0.822)
        };
        \addlegendentry{Drop}
      \end{axis}
    \end{tikzpicture}
}
\caption{Model performances tested on manipulated test data.}
\label{figure:ablation_study_on_keyword_manipulation_just_drop}
\end{centering}
\end{figure*}

We experiment with the PLMs and LLMs on \textsc{PARADISE} using the setup explained in Sec~\ref{sec:experimentsal_setup}. We also calculate the random and majority baselines, and evaluate the human performance by averaging the accuracy of two human annotators\footnote{Two university students majoring in computer science, between the ages of 20 - 24.} on a random set of 100 examples. Our main results are given in Table~\ref{table:inference_tasks}.

\paragraph{Fine-tuning} Among the fine-tuned models, DeBERTa performs the best in both tasks; yet, it still falls behind human performance. Considering DeBERTa's performance in previous abductive reasoning datasets, such as CosmosQA \citep{huang-etal-2019-cosmos}, SWAG \citep{zellers-etal-2018-swag}, and HellaSWAG \citep{zellers-etal-2019-hellaswag}, such results indicate that its success is not task-specific and its performance is transferable to warning and tip inference. DistilBERT, BERT, and RoBERTa cannot perform on par with DeBERTa and fall well behind human performance, although they perform considerably well on the proposed tasks. All models perform better in tip inference compared to warning inference.

\paragraph{Zero-shot} Although their performances are worse than fine-tuned PLMs, proprietary LLMs perform considerably better than open-source LLMs. GPT-4, which also tops many benchmarks \citep{openai2023gpt4}, is the best-performing LLM, while PALM-2 is a close runner-up. One surprising finding is that the performances of open-source LLMs are not correlated with their size, as Mistral 7B outperforms Vicuna 33B and LLaMA-2 70B in both tasks. Similar to PLMs, LLMs also perform better in tip inference compared to warning inference, with LLaMA-2 70B being an exception.

To gain further insights on behaviours of the models we ask the following research questions:
\begin{itemize}
\item \textbf{RQ1:} Do the models perform well due to simple keyword matching? 
\item \textbf{RQ2:} Do different model families fail on different instances? Is there a certain pattern?
\item \textbf{RQ3:} How does the performance compare for the explicit (i.e., directly related to a step) and implicit warnings/tips (i.e., more general and not directly related to any of the steps)?
\item \textbf{RQ4:} Are the models also good at the reverse task, i.e., can they find the goal most related to the warning/tip?
\item \textbf{RQ5:} Can the proposed tasks help improve performance in other procedural tasks?
\end{itemize}

\subsection{RQ1: Keyword Matching}
\label{subsection:keyword_manipulation}
\begin{figure*}[ht]
        \centering       \includegraphics[width=\linewidth]{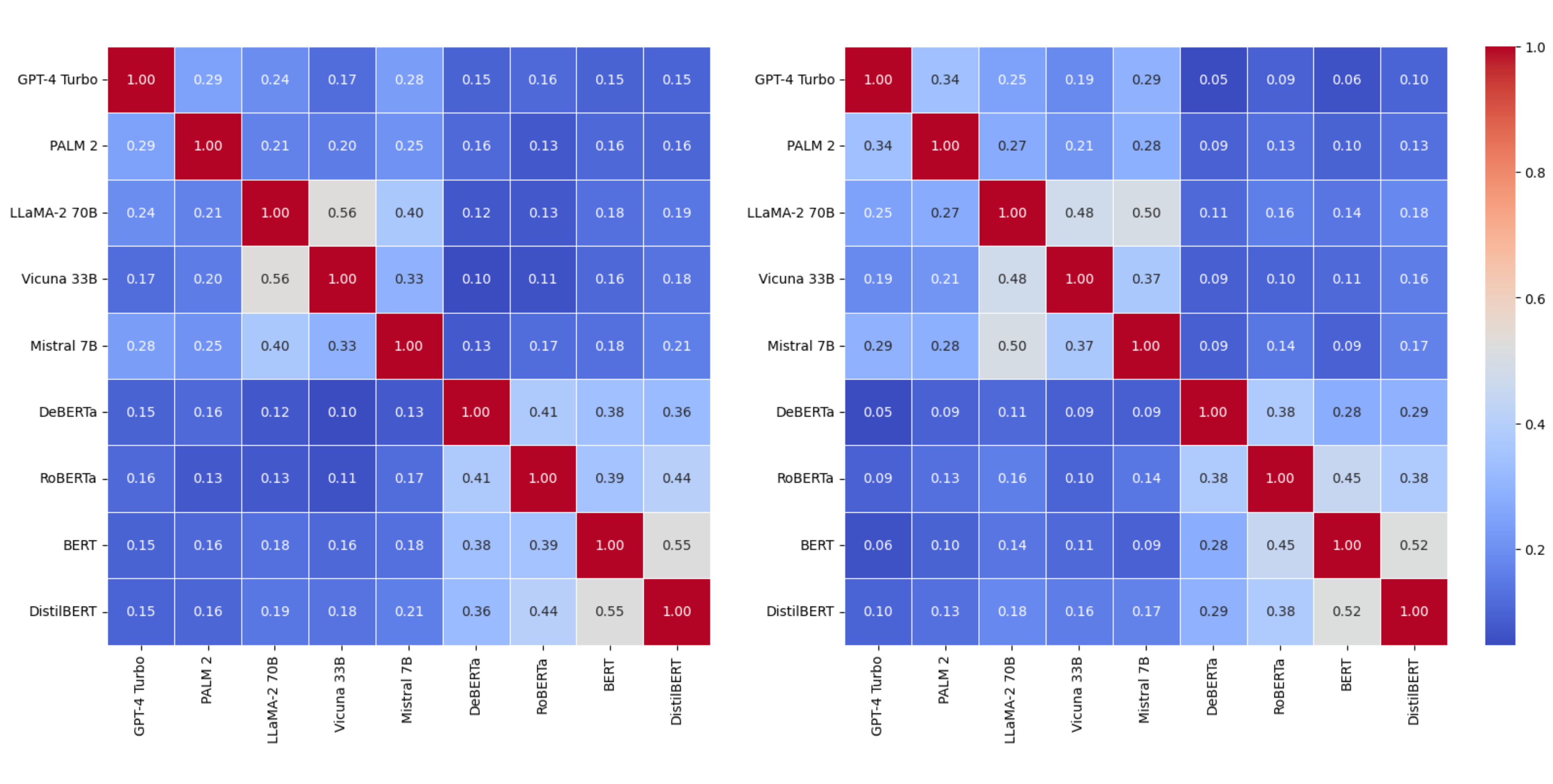}
        \caption{Correlation matrices of incorrect predictions of each model for tip (left) and warning (right) inference.}
        \label{fig:corr_matrices}
    \end{figure*}
If the goal and only one candidate share a common keyword, an example might become trivial and the task might develop into simple keyword matching. For example, when the goal and the positive candidate are about cats while negative candidates are about other animals, the positive candidate becomes easily distinguishable. Therefore, we drop such keywords in both positive and negative candidates in order to evaluate our tasks' dependency on keywords.

This manipulation, averaging approximately 2 tokens per candidate, induces a 4.5\% change on average. As illustrated in Fig. \ref{figure:ablation_study_on_keyword_manipulation_just_drop}, the omission of such keywords results in a 15\% to 20\% decrease in prediction accuracy for Pre-trained Language Models (PLMs). The impact diminishes with increasing original model performance; DistilBERT is most affected, while DeBERTa is least affected. In contrast, Language Models (LLMs) experience a milder accuracy decline of 5\% to 15\%. In addition to dropping keywords, we also experiment with other keyword manipulation methods as detailed in Appendix \ref{sec:additional_keyword_manipulation_methods}.

\subsection{RQ2: Failures of Different Model Families}
\label{subsection:plm_llm_comparison}
    To better understand the behaviours of the models we experiment with, we generate correlation matrices for their incorrect predictions in both tasks. As depicted in Fig.~\ref{fig:corr_matrices}, models within the same group (PLMs, Open-Source LLMs, and Proprietary LLMs) exhibit the highest correlation. Notably, incorrect predictions diverge more between PLMs and LLMs, while open-source LLMs and proprietary LLMs display higher inter-group correlation. Task-wise, correlation levels remain consistent, underscoring data quality and overall consistency.
    
    As evident in Fig.~\ref{fig:corr_matrices}, DeBERTa and GPT-4 exhibit divergent failure patterns. To understand their distinctions in distinguishing PLMs and LLMs, we manually inspect instances of failure. Our analysis reveals that DeBERTa struggles more with tangible, physical, and craft-related goals, while GPT-4 encounters challenges with abstract, digital, and social objectives. We validate these findings by generating the category distribution of unique failures for DeBERTa and GPT-4, as shown in Table \ref{table:category_distribution_of_failures}. Notably, GPT-4 falters in social and digital categories like Youth, Relationships, and Computer \& Electronics, while DeBERTa encounters difficulties in tangible categories such as Sports \& Fitness, Pets \& Animals, and Home \& Garden. Specific instances of failures for DeBERTa and GPT-4 are detailed in Appendix \ref{sec:deberta_gpt4_fails}.
    
    \begin{table}
        \begin{center}
        \scalebox{0.7}{
            \begin{tabular}{r|ccc|cc}
            \hline
            \multirow{2}{*}{\textbf{Rank}} & \multicolumn{3}{c|}{\textbf{DeBERTa}} & \multicolumn{2}{c}{\textbf{GPT-4}} \\
            \cline{2-6}
            & & \textbf{Warning} & \textbf{Tip} & \textbf{Warning} & \textbf{Tip} \\
            \toprule
            \textbf{\#1} & & \small{H\&G (24.5\%)} & \small{F\&E (25.0\%)} & \small{H\&G (24.4\%)} & \small{HE (18.1\%)}\\
            \textbf{\#2} & & \small{S\&F (12.8\%)} & \small{H\&G (21.9\%)} & \small{HE (16.3\%)} & \small{F\&E (13.4\%)} \\
            \textbf{\#3} & & \small{E\&C (11.7\%)} & \small{HE (15.6\%)}&  \small{C\&E (13.0\%)} & \small{C\&E (10.7\%)}\\
            \textbf{\#4} & & \small{PC\&S (11.7\%)} & \small{E\&C (10.9\%)} & \small{RE (9.8\%)} & \small{YO (8.7\%)} \\
            \textbf{\#5} & & \small{P\&A (8.5\%)} & \small{P\&A (9.4\%)} & \small{PC\&S (8.9\%)} & \small{PC\&S (8.7\%)} \\
            \bottomrule                \end{tabular}}
            \caption{Top categories that DeBERTa and GPT-4 fail.}
            \label{table:category_distribution_of_failures}
        \end{center}
    \end{table}

\subsection{RQ3: Implicit versus Explicit}
\label{subsection:matching_subsets}
    \begin{table*}
        \begin{center}
        \scalebox{0.9}{
            \begin{tabular}{lcccccccc}
                \hline
                 & & \multicolumn{3}{c}{\textbf{Warning}} & & \multicolumn{3}{c}{\textbf{Tip}} \\
                \cline{3-5} \cline{7-9}
                \textbf{Model} & & \textbf{All} & \textbf{Sim>0.5} & \textbf{Sim>0.75} & & \textbf{All} & \textbf{Sim>0.5} & \textbf{Sim>0.75} \\
                \toprule
                \multicolumn{9}{c}{\textbf{Fine-Tuned PLMs}} \\
                \midrule
                DistilBERT & & 82.16 \textpm 0.79 & \textbf{87.12 \textpm 0.66} & 85.71 \textpm 0.00 & & 87.48 \textpm 0.65 & 88.43 \textpm 0.67 & \textbf{90.02 \textpm 1.46} \\
                BERT & & 83.92 \textpm 0.68 & 86.84 \textpm 0.59 & \textbf{90.29 \textpm 2.29} & & 89.80 \textpm 0.91 & 91.10 \textpm 0.89 & \textbf{92.69 \textpm 0.77} \\
                RoBERTa & & 87.92 \textpm 0.60 & 89.57 \textpm 0.33 & \textbf{92.57 \textpm 1.40} & & 91.00 \textpm 0.34 & 91.83 \textpm 0.26 & \textbf{91.93 \textpm 0.77} \\
                DeBERTa & & 90.68 \textpm 0.41 & 91.63 \textpm 0.58 & \textbf{94.86 \textpm 2.14} & & 93.68 \textpm 0.48 & 95.34 \textpm 0.39 & \textbf{100.0 \textpm 0.0} \\
                \midrule
                \multicolumn{9}{c}{\textbf{Open-Source LLMs}} \\
                \midrule
                Mistral 7B & & 71.8 & 64.4 & \textbf{70.3} & & 72.4 & 72.8 & \textbf{73.6} \\
                Vicuna 33B & & 53.2 & 54.1 & \textbf{59.5} & & 57.0 & 58.9 & \textbf{66.0} \\
                LLaMA-2 70B & & \textbf{65.2} & 58.4 & 62.2 & & 64.5 & 70.9 & \textbf{77.4} \\
                \midrule
                \multicolumn{9}{c}{\textbf{Proprietary LLMs}} \\
                \midrule
                PALM-2 & & 83.6 & 83.2 & \textbf{86.5} & & 82.4 & 82.8 & \textbf{84.9} \\
                GPT-4 & & 86.2 & 86.1 & \textbf{89.2} & & 88.8 & 88.7 & \textbf{92.5} \\ 
                \bottomrule
            \end{tabular}}
        \captionsetup{justification=raggedright,singlelinecheck=false}
        \caption{The accuracy results of PLMs and LLMs on different subsets of the test splits with varying level of similarity to the instructions from associated wikiHow tutorials.}
        \label{table:matching_subsets}
        \end{center}
    \end{table*}
    As outlined in Sec.~\ref{sec:dataset}, warnings and tips within the dataset exhibit a distinction: some are specific to individual steps, while others are general. Although they are related to steps from the wikiHow tutorials, step-specific warnings and tips are not matched with their associated steps manually by editors. To assess the implicit reasoning skills of language models, we curate distinct subsets of test splits for both tasks, comprising warnings and tips demonstrating high semantic similarity with steps from relevant wikiHow tutorials. Employing SBERT \citep{reimers-gurevych-2019-sentence} for encoding steps and warnings/tips, we conduct a semantic similarity search. We retain examples with a cosine similarity score surpassing a threshold with at least one step, resulting in 35 warnings and 52 tips with high similarity scores (cosine similarity score with the step > 0.75) and 368 warnings and 382 tips with decent similarity scores (cosine similarity score with the step > 0.5).
    
    We assess the performance of PLMs and LLMs, as detailed in Sec.~\ref{sec:experimentsal_setup}, on subsets outlined in Table \ref{table:matching_subsets}. Results show improved model performance as the similarity between warnings/tips and steps increases. Higher accuracy in these subsets suggests a strong capacity for implicit reasoning, given that warnings and tips become more representative of intermediate steps with increased similarity. Notably, BERT and DeBERTa excel among PLMs, while Mistral 7B and LLaMA-2 70B lead among LLMs, exhibiting the highest accuracy increase in warning and tip inference tasks, respectively.

\subsection{RQ4: Reverse Inference Tasks}
If models can effectively reason about the relationship between warnings/tips and goals, it suggests they can correctly identify the goal corresponding to a given warning or tip. To examine this hypothesis, we construct reversed versions of our tasks, requiring the system to select the correct goal for a provided warning or tip. Using the candidate sampling method detailed in Sec \ref{subsection:candidate_sampling}, we randomly select 500 examples for evaluation with fine-tuned PLMs and LLMs. Results in Table \ref{table:reverse_tests} indicate zero-shot performances for PLMs align closely with the random baseline, except for DeBERTa, which achieves a 10\% higher accuracy. This suggests limited inherent reasoning capabilities over procedural warnings and tips. However, fine-tuned models exhibit a significant performance increase, supporting our hypothesis. In contrast, LLMs maintain similar accuracy scores in reverse tasks without experiencing a performance loss observed in fine-tuned PLMs.
\label{subsection:reverse_tasks}
    \begin{table}
        \begin{center}
        \scalebox{0.9}{
            \begin{tabular}{lccc}
                \hline
                {\textbf{Model}} & &
                    {\textbf{Reverse Warning}} &
                    {\textbf{Reverse Tip}} \\
                    \toprule
                    Random & & 25.0 & 25.0 \\
                    \midrule
                    \multicolumn{4}{c}{\textbf{PLMs}} \\
                    \midrule
                    DistilBERT &  & 20.52 \textpm 3.93 & 20.00 \textpm 5.27 \\
                    BERT & & 25.08 \textpm 3.73 & 25.40 \textpm 4.93 \\
                    RoBERTa & & 26.52 \textpm 5.42 & 26.36 \textpm 7.03 \\
                    \textbf{DeBERTa} & & \textbf{33.76 \textpm 4.83} & \textbf{35.72 \textpm 4.99} \\
                    \midrule
                    \multicolumn{4}{c}{\textbf{Fine-Tuned PLMs}} \\
                    \midrule
                    DistilBERT & & 65.44 \textpm 1.53 & 68.48 \textpm 1.25 \\
                    BERT & & 69.08 \textpm 1.69 & 72.36 \textpm 1.18 \\
                    RoBERTa & & 72.64 \textpm 1.44 & 77.00 \textpm 1.26 \\
                    \textbf{DeBERTa} & & \textbf{79.92 \textpm 0.63} & \textbf{80.44 \textpm 0.79} \\
                    \midrule
                    \multicolumn{4}{c}{\textbf{Open-Source LLMs}} \\
                    \midrule
                    \textbf{Mistral 7B} & & \textbf{74.6} & \textbf{79.2} \\
                    Vicuna 33B & &  62.8 & 61.4 \\
                    LLaMA-2 70B & & 76.2 & 76.0 \\
                    \midrule
                    \multicolumn{4}{c}{\textbf{Proprietary LLMs}} \\
                    \midrule
                    PALM-2 & & 83.6 & 85.8 \\
                    \textbf{GPT-4} & & \textbf{86.4} & \textbf{86.0} \\
                    \bottomrule
            \end{tabular}
            }
        \caption{Accuracy results of the reverse task evaluation.}
        \label{table:reverse_tests}
        \end{center}
    \end{table}
        
\subsection{RQ5: Transfer Learning}
\label{subsection:transfer_learning}
    
    To examine the impact of our tasks on reasoning over procedural documents, we conduct cross-tests between warning and tip inference tasks and out-of-domain transfer learning on goal and step inference tasks from \citet{zhang-etal-2020-reasoning}.
    
    \subsubsection{Cross Domain}
    \label{subsection:cross_tests}
        
        While categorized separately, both warnings and tips share the common objective of enhancing reader understanding in a wikiHow tutorial. As a result, they often exhibit similarities in structure and semantics, with occasional interchangeability. To assess the generalizability of Pre-trained Language Models (PLMs) across warning and tip inference tasks, we conduct cross-tests. Specifically, we evaluate PLMs fine-tuned on tip inference data using the test split of the warning inference dataset and vice versa.
        
        As depicted in Table \ref{table:cross_tests}, models fine-tuned on tip inference data demonstrate comparable performance to those fine-tuned on warning inference data for the warning inference task. Conversely, models fine-tuned on warning inference data exhibit slightly lower performance than those fine-tuned on tip inference data for the tip inference task. The nearly identical results in cross tests affirm the high similarity between warnings and tips, highlighting their interchangeability.
        \begin{table}
            \begin{center}
            \scalebox{1.0}{
                \begin{tabular}{lccc}
                    \hline
                    {\textbf{Model}} & &
                        {\textbf{Warning}} &
                        {\textbf{Tip}} \\
                        \toprule
                        Random & & 25.0 & 25.0 \\
                        \midrule
                        BERT & & 84.68 \textpm 0.27 & 86.20 \textpm 0.77 \\
                        RoBERTa & & 88.28 \textpm 0.30 & 88.80 \textpm 0.72 \\
                        \bottomrule
                \end{tabular}}
            \captionsetup{justification=raggedright,singlelinecheck=false}
            \caption{Accuracy results of BERT and RoBERta fine-tuned with tip inference on warning inference, and vice versa.}
            \label{table:cross_tests}
            \end{center}
        \end{table}
        
    \subsubsection{Out-of-Domain} 
        The goal and step inference tasks focus on identifying goal-step relationships in procedural how-to tutorials. \textit{Goal inference} involves selecting the plausible goal from candidates for a given step, while \textit{step inference} entails the reverse process. Although similar to the proposed tasks, they aim to measuring \textit{explicit} procedural reasoning abilities. 
        
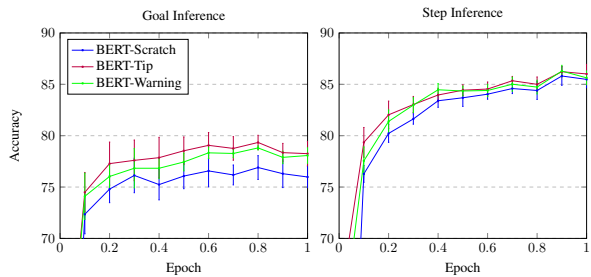
\begin{figure}[ht]
\begin{centering}
\resizebox{0.475\textwidth}{!}{%
    \begin{tikzpicture}
        \begin{axis}[
        title={Goal Inference},
        xlabel={Epoch},
        ylabel={Accuracy},
        xmin=0, xmax=1.0,
        ymin=70, ymax=90,
        xtick={0.0, 0.2, 0.4, 0.6, 0.8, 1.0},
        ytick={70, 75, 80, 85, 90},
        legend pos=north west,
        legend cell align=left,
        ymajorgrids=true,
        grid style=dashed,
    ]
    
    \addplot[
        color=blue,
        mark=o,
        mark size=0.5,
        error bars/.cd,
        y dir=both, 
        y explicit,
        ]
        coordinates {
        (0.00, 26.19) +- (0, 5.49)
        (0.10, 72.35) +- (0, 1.90)
        (0.20, 74.80) +- (0, 1.32)
        (0.30, 76.13) +- (0, 1.68)
        (0.40, 75.24) +- (0, 1.49)
        (0.50, 76.07) +- (0, 1.23)
        (0.60, 76.57) +- (0, 1.55)
        (0.70, 76.18) +- (0, 0.97)
        (0.80, 76.90) +- (0, 1.16)
        (0.90, 76.30) +- (0, 1.35)
        (1.00, 75.98) +- (0, 1.04)
        };
        \addlegendentry{BERT-Scratch}

    \addplot[
        color=purple,
        mark=o,
        mark size=0.5,
        error bars/.cd,
        y dir=both, 
        y explicit,
        ]
        coordinates {
        (0.00, 49.17) +- (0, 2.52)
        (0.10, 74.49) +- (0, 1.93)
        (0.20, 77.28) +- (0, 2.10)
        (0.30, 77.60) +- (0, 1.98)
        (0.40, 77.86) +- (0, 1.98)
        (0.50, 78.53) +- (0, 1.35)
        (0.60, 79.06) +- (0, 1.25)
        (0.70, 78.76) +- (0, 1.15)
        (0.80, 79.34) +- (0, 0.70)
        (0.90, 78.35) +- (0, 0.90)
        (1.00, 78.25) +- (0, 1.18)
        };
        \addlegendentry{BERT-Tip}

    \addplot[
        color=green,
        mark=o,
        mark size=0.5,
        error bars/.cd,
        y dir=both, 
        y explicit,
        ]
        coordinates {
        (0.00, 47.35) +- (0, 2.17)
        (0.10, 74.10) +- (0, 2.25)
        (0.20, 76.03) +- (0, 1.13)
        (0.30, 76.83) +- (0, 1.91)
        (0.40, 76.83) +- (0, 1.02)
        (0.50, 77.44) +- (0, 0.87)
        (0.60, 78.33) +- (0, 0.55)
        (0.70, 78.27) +- (0, 0.41)
        (0.80, 78.81) +- (0, 0.2)
        (0.90, 77.89) +- (0, 0.61)
        (1.00, 78.07) +- (0, 0.75)
        };
        \addlegendentry{BERT-Warning}
    \end{axis}
    \end{tikzpicture}%
    \begin{tikzpicture}
        \begin{axis}[
        title={Step Inference},
        xlabel={Epoch},
        ylabel={},
        xmin=0, xmax=1.0,
        ymin=70, ymax=90,
        xtick={0.0, 0.2, 0.4, 0.6, 0.8, 1.0},
        ytick={70, 75, 80, 85, 90},
        legend pos=north west,
        legend cell align=left,
        ymajorgrids=true,
        grid style=dashed,
    ]
    
    \addplot[
        color=blue,
        mark=o,
        mark size=0.5,
        error bars/.cd,
        y dir=both, 
        y explicit,
        ]
        coordinates {
        (0.00, 23.78) +- (0, 1.99)
        (0.10, 76.28) +- (0, 0.82)
        (0.20, 80.22) +- (0, 0.87)
        (0.30, 81.63) +- (0, 0.51)
        (0.40, 83.40) +- (0, 0.64)
        (0.50, 83.69) +- (0, 0.84)
        (0.60, 84.04) +- (0, 0.49)
        (0.70, 84.59) +- (0, 0.49)
        (0.80, 84.40) +- (0, 0.87)
        (0.90, 85.81) +- (0, 0.91)
        (1.00, 85.48) +- (0, 0.81)
        };

    \addplot[
        color=purple,
        mark=o,
        mark size=0.5,
        error bars/.cd,
        y dir=both, 
        y explicit,
        ]
        coordinates {
        (0.00, 63.55) +- (0, 1.01)
        (0.10, 79.35) +- (0, 1.45)
        (0.20, 82.03) +- (0, 1.34)
        (0.30, 83.03) +- (0, 0.79)
        (0.40, 83.96) +- (0, 0.66)
        (0.50, 84.43) +- (0, 0.53)
        (0.60, 84.54) +- (0, 0.69)
        (0.70, 85.35) +- (0, 0.42)
        (0.80, 85.00) +- (0, 0.72)
        (0.90, 86.23) +- (0, 0.51)
        (1.00, 86.01) +- (0, 0.93)
        };

    \addplot[
        color=green,
        mark=o,
        mark size=0.5,
        error bars/.cd,
        y dir=both, 
        y explicit,
        ]
        coordinates {
        (0.00, 58.53) +- (0, 1.58)
        (0.10, 77.55) +- (0, 1.36)
        (0.20, 81.39) +- (0, 1.12)
        (0.30, 82.98) +- (0, 0.72)
        (0.40, 84.47) +- (0, 0.60)
        (0.50, 84.35) +- (0, 0.48)
        (0.60, 84.40) +- (0, 0.60)
        (0.70, 85.01) +- (0, 0.68)
        (0.80, 84.73) +- (0, 0.67)
        (0.90, 86.29) +- (0, 0.56)
        (1.00, 85.63) +- (0, 0.67)
        };
    \end{axis}
    \end{tikzpicture}%
    }
\captionsetup{justification=raggedright,singlelinecheck=false}
\caption{Accuracy of the three BERT variations plotted over the training epoch.}
\label{fig:ablation_study_on_transfer_learning}
\end{centering}
\end{figure}

        For both step and goal inference tasks, we fine-tune three BERT models: i) BERT trained from scratch, ii) BERT previously fine-tuned on warning inference data, and iii) BERT previously fine-tuned on tip inference data. We report their performances on the test split throughout the training. Notably, prior fine-tuning on warning and tip inference tasks consistently improves performance during training for both goal and step inference tasks, as illustrated in Fig. \ref{fig:ablation_study_on_transfer_learning}. The second and third BERT models exhibit significantly enhanced zero-shot performances, with average accuracy increases of 21.16\% and 22.98\% for goal inference, and 34.75\% and 39.77\% for step inference, respectively. While the performance gap diminishes, the second and third BERT models continue to outperform at the end of training, with average accuracy increases of 2.09\% and 2.27\% for goal inference, and 0.15\% and 0.53\% for step inference, respectively.

\section{Related Work}
\label{sec:related_work}
Commonsense reasoning is a broad domain that branches into a wide range of subdomains such as linguistic reasoning \citep{sahin-etal-2020-linspector, liu-etal-2022-testing, lin-etal-2021-riddlesense}, abductive reasoning \citep{tandon-etal-2019-wiqa, zellers-etal-2019-hellaswag}, reasoning about the physical world \citep{Bisk2020, Khot2019QASCAD, aroca-ouellette-etal-2021-prost}, temporal reasoning \citep{zhang-etal-2020-reasoning, qin-etal-2021-timedial}, etc. \citep{Bhargava2022CommonsenseKR}. Although there exists some abductive reasoning tasks that involve finding the most likely explanation for a set of incomplete observations \citep{Bhargava2022CommonsenseKR}, they are outside of the domain of procedural language understanding. For example, CosmosQA \citep{huang-etal-2019-cosmos} present commonsense abductive reading comprehension and ART \citep{bhagavatula2020abductive} propose abductive natural language inference with narrative contexts. Thus, they depend on additional paragraphs provided by the question, which enrich the level of information provided to the model. Furthermore, abductive reasoning resources obtained from procedural texts are either artificially made difficult with targeted models (such as SWAG \citep{zellers-etal-2018-swag} and HellaSWAG \citep{zellers-etal-2019-hellaswag}) or covers another form of procedural texts (e.g., natural phenomenons of WIQA \citep{tandon-etal-2019-wiqa}). Moreover, they focus on the continuous elements (i.e. consecutive steps or events), which we believe diminishes the degree of implicity. 

Furthermore our main resource wikiHow has been extensively used for a wide range of tasks thanks to its rich body of well-structured procedural documents, including but not limited to summarization \citep{Koupaee2018WikiHowAL, ladhak-etal-2020-wikilingua}, intent detection \citep{zhang-etal-2020-intent}, reasoning \citep{zhang-etal-2020-reasoning}, linking actions \citep{lin-etal-2020-recipe, zhou-etal-2022-show}, and next event prediction \citep{nguyen-etal-2017-sequence, zellers-etal-2018-swag, zellers-etal-2019-hellaswag}.

\section{Conclusion}
\label{sec:conclusion}
There has been a growing interest in procedural data, tasks, and reasoning. However, the spotlight has been on explicit and direct relations when studying reasoning within procedural documents. To address the the lack of resources to study implicit relations and reasoning, we introduce \texttt{PARADISE} and strong baseline models evaluated and analyzed with extensive experiments. \texttt{PARADISE} contains +104K warnings and tips in total and serves as a reliable testbed for the evaluation of abductive and implicit commonsense reasoning skills of language models. Moreover, it brings improvement to zero-and-few-shot performances in out-of-domain procedural tasks. Our experiments reveal that PLMs do not possess inherent reasoning skills; yet, most of them outperform LLMs when fine-tuned. However, best-performing models from both groups fall behind human performance, indicating room for improvement. We release all the resources publicly to further research.


\section*{Limitations}
    We evaluate LLMs with their respective APIs due to their proprietary nature or heavy computation costs. Therefore, although we explain our evaluation setup in detail, the performances of LLMs we evaluate might not always be reproducible due to potential future changes or deprecations of their APIs. 

\section*{Ethics Statement}
    We utilize the content from wikiHow, adhering to the specific circumstances outlined in the Creative Commons license. We fully comply with all conditions stipulated by the Creative Commons license, and these requirements facilitate the utilization of the wikiHow corpus upon which we build.

\section*{Acknowledgements}
    This work has been supported by the Scientific and Technological Research Council of Türkiye~(TÜBİTAK) as part of the project ``Automatic Learning of Procedural Language from Natural Language Instructions for Intelligent Assistance'' with the number 121C132. We also gratefully acknowledge KUIS AI Lab for providing computational support. We thank our anonymous reviewers and the members of GGLab who helped us improve this paper. We especially thank Aysha Gurbanova, Şebnem Demirtaş, and Mahmut İbrahim Deniz for their contributions to evaluating human performance on warning and tip inference tasks.

\bibliography{anthology,custom}

\appendix

\section{Example JSON File}
\lstdefinestyle{json}{
    language=json,
    basicstyle=\small\ttfamily,
    commentstyle=\color{gray},
    stringstyle=\color{blue},
    keywordstyle=\color{red}
}
\colorlet{punct}{red!60!black}
\definecolor{background}{HTML}{EEEEEE}
\definecolor{delim}{RGB}{20,105,176}
\colorlet{numb}{magenta!60!black}

An example JSON file from our corpus can be seen in Listing \ref{lst:json-example}.

\lstdefinelanguage{json}{
    basicstyle=\tiny\ttfamily,
    numbers=left,
    numberstyle=\scriptsize,
    stepnumber=1,
    numbersep=8pt,
    showstringspaces=false,
    breaklines=true,
    frame=lines,
    backgroundcolor=\color{background},
    literate=
     *{0}{{{\color{numb}0}}}{1}
      {1}{{{\color{numb}1}}}{1}
      {2}{{{\color{numb}2}}}{1}
      {3}{{{\color{numb}3}}}{1}
      {4}{{{\color{numb}4}}}{1}
      {5}{{{\color{numb}5}}}{1}
      {6}{{{\color{numb}6}}}{1}
      {7}{{{\color{numb}7}}}{1}
      {8}{{{\color{numb}8}}}{1}
      {9}{{{\color{numb}9}}}{1}
      {:}{{{\color{punct}{:}}}}{1}
      {,}{{{\color{punct}{,}}}}{1}
      {\{}{{{\color{delim}{\{}}}}{1}
      {\}}{{{\color{delim}{\}}}}}{1}
      {[}{{{\color{delim}{[}}}}{1}
      {]}{{{\color{delim}{]}}}}{1},
}
\begin{lstlisting}[
    style=json,
    language=json,
    caption={Example JSON file for the procedural tutorial with the goal "Rip Your Own Jeans" from wikiHow.},
    label={lst:json-example}
]
{
    "title": "How to Rip Your Own Jeans",
    "url": "https://www.wikihow.com/Rip-Your-Own-Jeans",
    "title_description": "Distressed denim is a popular style, but buying jeans that are already ripped can be expensive. Luckily, you can create this trend yourself by roughing up the fabric with a piece of sandpaper, then snipping a hole with a pair of scissors.",
    "category_hierarchy": [
        "Hobbies and Crafts",
        "Crafts",
        "Decoration Projects"
    ],
    "author": {
        "name": null,
        "n_coauthors": 128,
        "blurb": null
    },
    "time_updated": "April 12, 2020",
    "n_views": 3443113,
    "rating": {
        "n_votes": 76,
        "helpful_percent": 77
    },
    "methods": [],
    "parts": [],
    "video": "https://www.wikihow.com/Video/Rip-Your-Own-Jeans",
    "related_articles": ["ABRIDGED"],
    "tips": [
        "Washing the jeans right after ripping them causes the fibers to loosen more and create a more distressed look.",
        "Avoid adding rips too near the seams, as they may cause them to begin to unravel.",
        "For exact rips, use a sewing needle to pull out individual stitches from the fabric."
    ],
    "warnings": [
        "Don't make the rip too big at first. Washing the fabric will increase the size and fray of the hole.",
        "Never attempt to rip or fray your jeans while you're wearing them.",
        "Use caution with sharp tools."
    ],
    "QAs": ["ABRIDGED"],
    "refs": [
        "https://www.marieclaire.co.uk/news/fashion-news/how-to-rip-jeans-821",
        "https://www.marieclaire.co.uk/news/fashion-news/how-to-rip-jeans-821",
        "https://www.cosmopolitan.com/style-beauty/fashion/a58592/5-easy-tricks-for-distressing-your-jeans/"
    ],
    "quizzes": [],
    "other_languages": {"ABRIDGED"},
    "steps": ["ABRIDGED"]
}
\end{lstlisting}
\label{sec:example_json_file}

\section{Filtering Examples}
After sampling the negative candidates, we apply a set of filters introduced by \citet{zhang-etal-2020-reasoning} to ensure the high-quality of the pairs and the challenge they bring. However we change some of these filters as follows:

\paragraph{Similarity filter:} We obtain the cosine similarity scores using Supervised SimCSE-RoBERTa-Large \citep{gao-etal-2021-simcse}, since sentence embeddings capture sentence-level information, resulting in better filtering.

\paragraph{Length filter:} We also set an upper bound to ensure they are long ($>8$ tokens) enough to contain relevant information yet short ($<128$ tokens) enough to be on-point and coherent.

\paragraph{Category filter:} We exclude examples from some categories (e.g., Philosophy and Religion, Celebrities, Holidays and Traditions, etc.) that might not be suitable for evaluating language models' reasoning skills due to their subjectivity.
\label{sec:filtering_examples}

\section{Dataset Analysis with Cartography}
\begin{figure*}%
    \centering
   \includegraphics[scale=0.21]{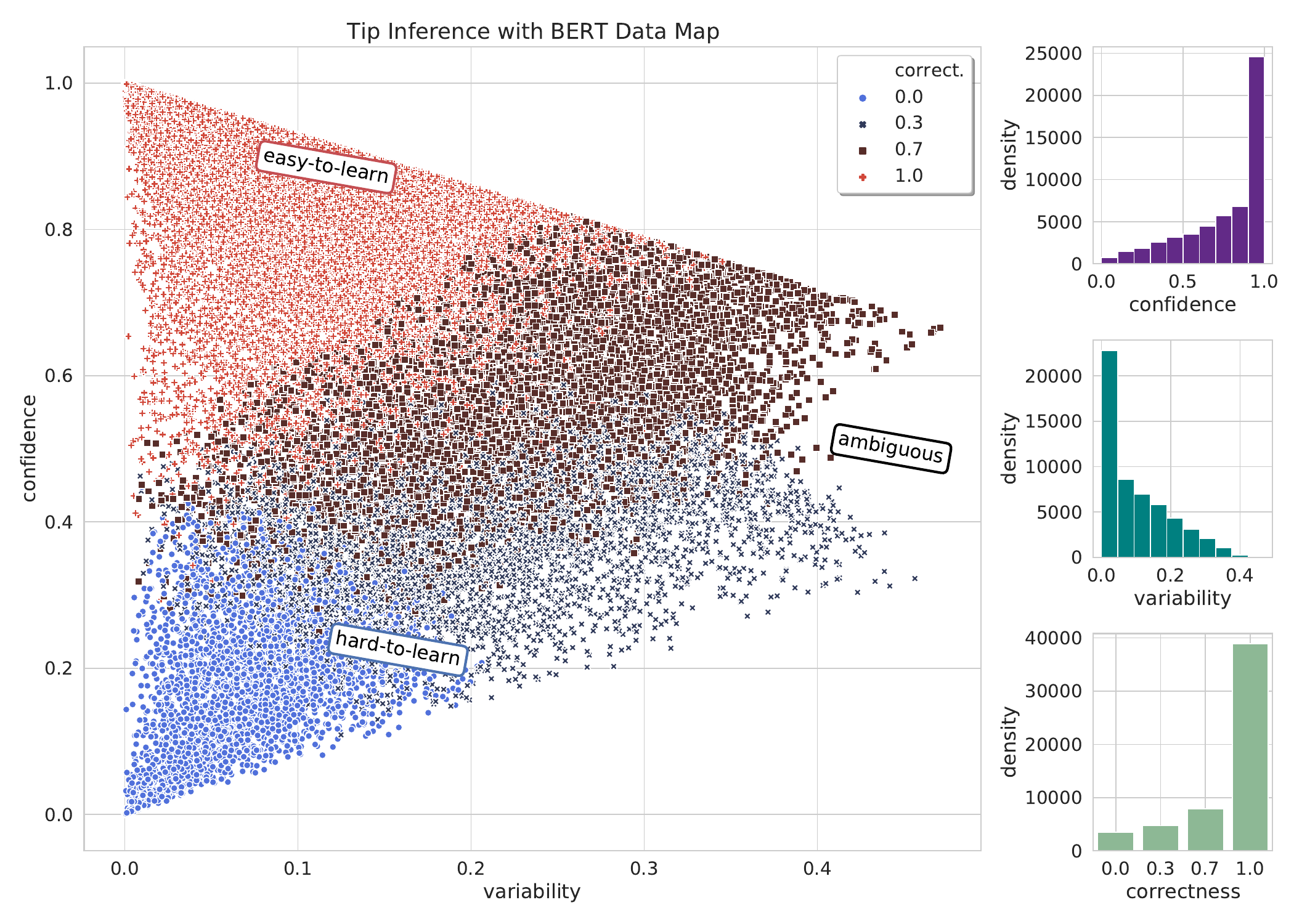}%
    \qquad
    \includegraphics[scale=0.21]{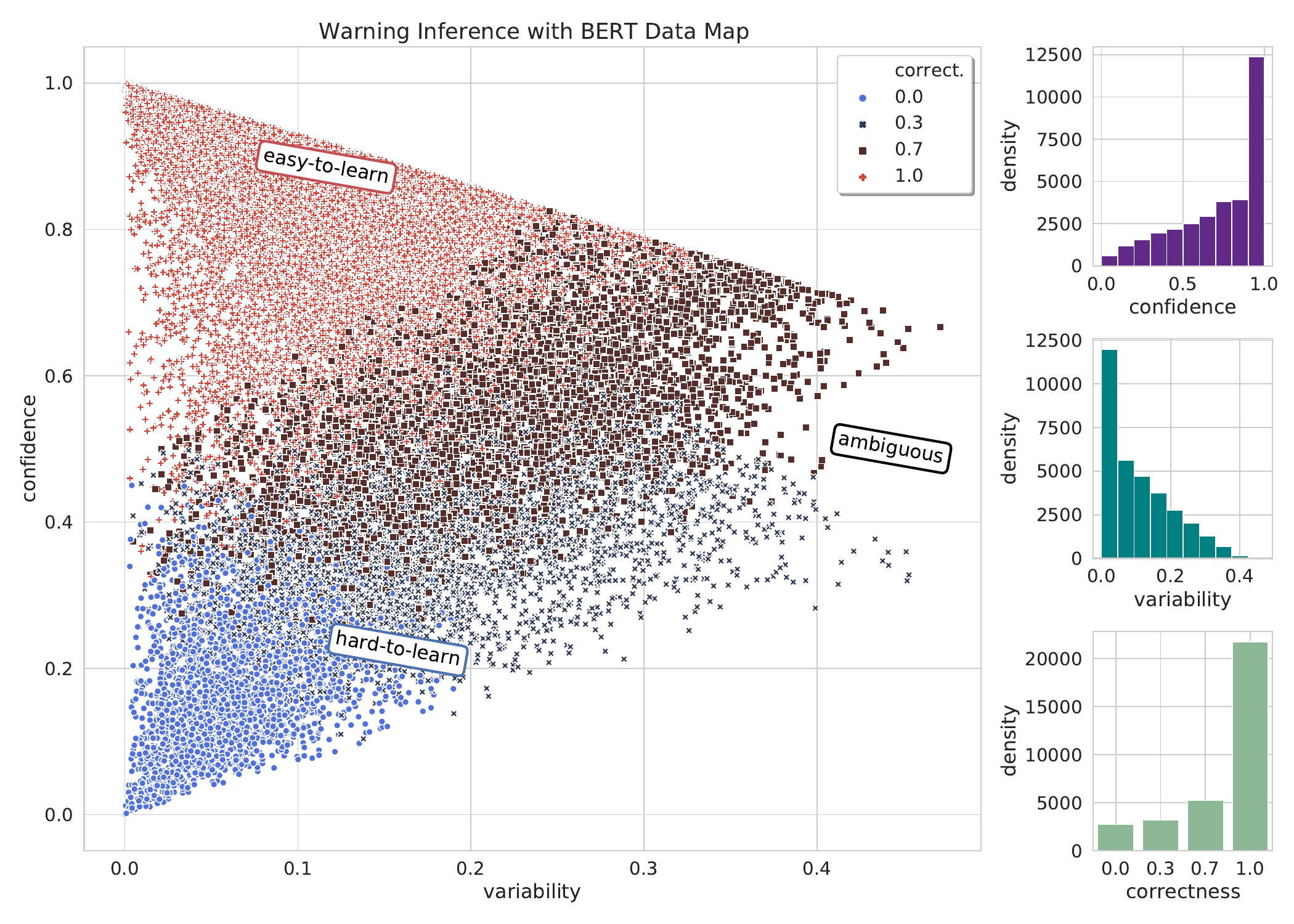}%
    \caption{Data maps for warning and tip inference tasks obtained with BERT.}%
    \label{fig:data_maps}%
\end{figure*}

We utilize the dataset cartography tool \citep{swayamdipta-etal-2020-dataset}, which processes a model's behaviour on training instances (also known as the training dynamics) for mapping the dataset, to better understand the characteristics of our datasets. To this end, cartography derives three metrics from the training dynamics: confidence (the mean model probability of the true label across epochs), correctness (the fraction of times the model correctly labels an observation across epochs), and variability (the spread of the model’s probability of correctly labeling observations across epochs). With these metrics, cartography reveals three regions in a data map, each with distinct features.

Using the cartography tool, we generate data maps for warning and tip inference datasets. As seen in Fig. \ref{fig:data_maps}, both of our datasets have a high density in the positive ends of confidence and correctness and in the negative end of variability, indicating that BERT is capable of confidently choosing the correct warnings and tips throughout the training. However, tip inference dataset shows greater density in those ends, demonstrating higher easiness that reinforces our reasoning in Appendix~\ref{sec:warnings_vs_tips}.

Additionally, we implement the noise detector \citep{swayamdipta-etal-2020-dataset} using a Gaussian Naive Bayes classifier model \citep{gnb} to identify mislabeled instances in our datasets. We train our classifier model on a small set of equally distributed mislabeled (randomly re-assinged) and correctly labeled data instances. Although simple, it achieves 95.2\% accuracy on the test set. We, then, use the classifier model on the entire warning and tip inference datasets. Our classifier model finds 1022 noisy instances in the tip inference and 746 noisy instances in the warning inference datasets, indicating a 1.3\% and 1.9\% of noise respectively. Such levels of noise in automatically generated datasets with no human supervision (other than curating the test splits) illustrates the success of the filtering described in Appendix \ref{sec:filtering_examples}.

\label{sec:dataset_analysis_with_cartography}

\section{Implementation Details}
We use the base versions of DistilBERT\footnote{\url{https://huggingface.co/distilbert-base-uncased}}, BERT\footnote{\url{https://huggingface.co/bert-base-uncased}}, RoBERTa\footnote{\url{https://huggingface.co/roberta-base}}, and DeBERTa\footnote{\url{https://huggingface.co/microsoft/deberta-v3-base}} and implement them using Hugging Face libraries, namely transformers \citep{Wolf_Transformers_State-of-the-Art_Natural_2020}, evaluate, and accelerate \citep{accelerate}.

We fine-tune each model for 1 epoch with its unique set of hyperparameters that can be seen in Table \ref{table:hyperparameters}. We use a batch size of 8 and keep the default values for all other hyperparameters for testing. In fine-tuning, we use five different seeds (42, 2717, 6802, 9893, and 7818) to conduct statistical significance analysis. In testing, we set the seed to 42. 

\begin{table}
    \begin{center}
    \scalebox{0.65}{
    \begin{tabular}{lcccc}
        \textbf{Hyperparameter} & \textbf{DistilBERT} & \textbf{BERT} & \textbf{RoBERTa} & \textbf{DeBERTa} \\
        \toprule
        Total Batch Size & 32 & 64 & 64 & 128 \\
        Gradient Acc. Steps & 1 & 2 & 2 & 4 \\
        Learning Rate & 2e-5 & 2e-5 & 1e-5 & 1e-5 \\
        Max. Sequence Length & 128 & 128 & 128 & 128 \\
        \bottomrule
    \end{tabular}}
    \captionsetup{justification=raggedright,singlelinecheck=false}
    \caption{Hyperparameters used in the fine-tunings of models.}
    \label{table:hyperparameters}
    \end{center}
\end{table}

For both fine-tuning and testing, we utilize half-precision floating point format (FP16) with the accelerate library. DistilBERT, BERT, and RoBERTa models are fine-tuned on a single NVIDIA T4, while DeBERTa is fine-tuned on two NVIDIA T4s. Computational costs of fine-tuning each model across each task can be seen in Table \ref{table:computational_costs}.

\begin{table}
    \begin{centering}
    \scalebox{0.75}{
    \begin{tabular}{lcc}
        \textbf{Model} & \textbf{GPU} & \textbf{Time} \\
        \toprule
        \textsc{Warning Inference} & & \\
        DistilBERT & 1 x NVIDIA T4 & 6 mins 10 secs \\
        BERT & 1 x NVIDIA T4 & 11 mins 53 secs \\
        RoBERTa & 1 x NVIDIA T4 & 11 mins 45 secs \\
        DeBERTa & 2 x NVIDIA T4 & 9 mins 32 secs \\
        \midrule
        \textsc{Tip Inference} & & \\
        DistilBERT & 1 x NVIDIA T4 & 12 mins 47 secs \\
        BERT & 1 x NVIDIA T4 & 24 mins 57 secs \\
        RoBERTa & 1 x NVIDIA T4 & 24 mins 51 secs \\
        DeBERTa & 2 x NVIDIA T4 & 19 mins 55 secs \\
        \bottomrule
    \end{tabular}}
    \captionsetup{justification=raggedright,singlelinecheck=false}
    \caption{Computational costs of fine-tuning each model across each task.}
    \label{table:computational_costs}
    \end{centering}
\end{table}
\label{sec:implementation_details}

\section{Prompting LLMs}
In order to effectively engage with a language model, it is essential to meticulously construct a prompt template and fine\-tune specific parameters, notably temperature (sampling temperature between 0 and 1 or 0 and 2) and top p (nucleus sampling, where the model considers the results of the tokens with top p probability mass)~\cite{OpenAI2023api}.

For the prompt template construction, we referred to the official API documentation of each model to ascertain the recommended templates. In cases where no specific templates were provided, we directly used the questions as prompts, omitting any additional tokens.

Regarding the calibration of the temperature and top p settings, we initiated our tests with the default values as specified in the model's API and playground interface. This approach was followed by a systematic tuning process to optimize performance. Notably, we observed that a lower temperature setting, as compared to the default, yielded more accurate results. This aligns with the general understanding that lower temperatures are preferable for fact-based prompts, while higher temperatures are better suited for tasks requiring creativity and an element of randomness.

The accompanying tables \ref{tab:llms_tips_prompts} and \ref{tab:llms_warns_prompts} show the specific prompts and parameter settings employed for each model (all the experiments were done in December 2023).

\begin{table*}[htbp]
    \centering
    \begin{tabular}{|m{1.5cm}|m{2cm}|m{13cm}|}
        \hline
        \textbf{Model} & \textbf{Settings} & \textbf{Prompt} \\ 
        \hline
        \makecell{\textbf{GPT-4} \\ \\ \\ \textbf{PALM2}} & \makecell{\textbf{Temp. = 0.3} \\ \textbf{Top p = 0.9} \\ \\ \textbf{Temp. = 0.3} \\ \textbf{Top p = 0.9}} & I will give you a goal below and 4 tips. Can you pick the most related tip to it?\newline
        \textbf{Goal}: Prepare for a Long Car Trip\newline
        \textbf{Tips}:\newline
        \textbf{Tip 0-} While performing any exercise, make sure you are drinking water to stay hydrated.\newline
        \textbf{Tip 1-} Do drink plenty of water to keep your skin hydrated.\newline
        \textbf{Tip 2-} If you are traveling for a long time, bring a bottle of water to keep you hydrated.\newline
        \textbf{Tip 3-} Bring a bottle of water with you to stay hydrated.\newline
        \textbf{Response format}: Return the tip number in json with `tip' as key and no more details. \\
        \hline
        
        \makecell{\textbf{Llama2} \\ \\ \\ \textbf{Mistral}} & \makecell{\textbf{Temp. = 0.3} \\ \textbf{Top p = 0.9} \\ \\ \textbf{Temp. = 0.0} \\ \textbf{Top p = 0.1}} & 
        \sloppy{<s>[INST]\newline
        I will give you a goal below and 4 tips. Can you pick the most related tip to it?\newline
        \textbf{Goal}: Prepare for a Long Car Trip\newline
        \textbf{Tips}:\newline
        \textbf{Tip 0-} While performing any exercise, make sure you are drinking water to stay hydrated.\newline
        \textbf{Tip 1-} Do drink plenty of water to keep your skin hydrated.\newline
        \textbf{Tip 2-} If you are traveling for a long time, bring a bottle of water to keep you hydrated.\newline
        \textbf{Tip 3-} Bring a bottle of water with you to stay hydrated.\newline
        \textbf{Response format}: Return the tip number in json with `tip' as key and no more details. \newline
        [$\backslash$/INST]}
        \\
        \hline
        
        \makecell{\textbf{Vicuna}} & \makecell{\textbf{Temp. = 0.3} \\ \textbf{Top-p = 0.9}} & 
        \sloppy{A chat between a human and an assistant.\newline
        $\#\#\#$ Human:\newline
        \textbf{Goal}: Prepare for a Long Car Trip\newline
        \textbf{Tips}:\newline
        \textbf{Tip 0-} While performing any exercise, make sure you are drinking water to stay hydrated.\newline
        \textbf{Tip 1-} Do drink plenty of water to keep your skin hydrated.\newline
        \textbf{Tip 2-} If you are traveling for a long time, bring a bottle of water to keep you hydrated.\newline
        \textbf{Tip 3-} Bring a bottle of water with you to stay hydrated.\newline
        \textbf{Response format}: Return the tip number in json with `tip' as key and no more details.\newline 
        $\#\#\#$ Assistance:}
        \\
        \hline
    \end{tabular}
    \caption{LLM settings and prompts for tip inference.}
    \label{tab:llms_tips_prompts}
\end{table*}

\begin{table*}[htbp]
    \centering
    \begin{tabular}{|m{1.5cm}|m{2cm}|m{13cm}|}
        \hline
        \textbf{Model} & \textbf{Settings} & \textbf{Prompt} \\ 
        \hline
        \makecell{\textbf{GPT-4} \\ \\ \\ \textbf{PALM2}} & \makecell{\textbf{Temp. = 0.3} \\ \textbf{Top p = 0.9} \\ \\ \textbf{Temp. = 0.3} \\ \textbf{Top p = 0.9}} & I will give you a goal below and 4 warnings. Can you pick the most related warning to it?\newline
        \textbf{Goal}: Make Laundry Detergent Slime\newline
        \textbf{Warnings}:\newline
        \textbf{Warn 0-} Warning 0- Avoid flooding the floor with cleaner or water. A thin layer of water should be enough for a dry cloth to wipe\newline
        \textbf{Warn 1-} Don't apply heat (dryer, iron) to the stained area until the stain is gone.\newline
        \textbf{Warn 2-} Don't place the slime in a cold area when it's finished. It may become less stretchy.\newline
        \textbf{Warn 3-} Don't let the resurfacer dry on your skin since it may cause irritation and is difficult to remove. \newline
        \textbf{Response format}: Return the warning number in json with `warn' as key and no more details. \\
        \hline
        
        \makecell{\textbf{Llama2} \\ \\ \\ \textbf{Mistral}} & \makecell{\textbf{Temp. = 0.3} \\ \textbf{Top p = 0.9} \\ \\ \textbf{Temp. = 0.0} \\ \textbf{Top p = 0.1}} & 
        \sloppy{<s>[INST]\newline
        I will give you a goal below and 4 warnings. Can you pick the most related warning to it?\newline
        \textbf{Goal}: Make Laundry Detergent Slime\newline
        \textbf{Warnings}:\newline
        \textbf{Warn 0-} Warning 0- Avoid flooding the floor with cleaner or water. A thin layer of water should be enough for a dry cloth to wipe\newline
        \textbf{Warn 1-} Don't apply heat (dryer, iron) to the stained area until the stain is gone.\newline
        \textbf{Warn 2-} Don't place the slime in a cold area when it's finished. It may become less stretchy.\newline
        \textbf{Warn 3-} Don't let the resurfacer dry on your skin since it may cause irritation and is difficult to remove. \newline
        \textbf{Response format}: Format the answer in json with the warning number as value and 'warn' as key and no more details.
        [$\backslash$/INST]}
        \\
        \hline
        
        \makecell{\textbf{Vicuna}} & \makecell{\textbf{Temp. = 0.3} \\ \textbf{Top-p = 0.9}} & 
        \sloppy{A chat between a human and an assistant.\newline
        $\#\#\#$ Human:\newline
          I will give you a goal below and 4 warnings. Can you pick the most related warning to it?\newline
            \textbf{Goal}: Make Laundry Detergent Slime\newline
            \textbf{Warnings}:\newline
            \textbf{Warn 0-} Warning 0- Avoid flooding the floor with cleaner or water. A thin layer of water should be enough for a dry cloth to wipe\newline
            \textbf{Warn 1-} Don't apply heat (dryer, iron) to the stained area until the stain is gone.\newline
            \textbf{Warn 2-} Don't place the slime in a cold area when it's finished. It may become less stretchy.\newline
            \textbf{Warn 3-} Don't let the resurfacer dry on your skin since it may cause irritation and is difficult to remove. \newline
            \textbf{Response format}: Return the warning number in json with `warn' as key and no more details.
            }
        \\
        \hline
    \end{tabular}
    \caption{LLM settings and prompts for warning inference.}
    \label{tab:llms_warns_prompts}
\end{table*}
\label{sec:prompting_llms}

\section{Additional Keyword Manipulation Methods}
\begin{figure*}[ht]
\begin{centering}
\resizebox{0.85\textwidth}{!}{%
    \begin{tikzpicture}
        \begin{axis}[
        title={Warning Inference},
        xlabel={},
        ylabel={Accuracy},
        ybar=1.0pt,
        bar width=4.5pt,
        symbolic x coords={DistilBERT, BERT, RoBERTa, DeBERTa},
        grid=both,
        ymin=0,
        legend cell align=left,
        enlarge x limits=0.15,
        ymin=0.5,
        ymax=1,
        x label style={font=\footnotesize},
        y label style={font=\footnotesize},
        ticklabel style={font=\footnotesize},
        ]
        \addplot[black,fill=red,error bars/y dir=both,error bars/y explicit] coordinates {
          (DistilBERT, 0.8216) +- (0, 0.0079)
          (BERT, 0.8392) +- (0, 0.0068)
          (RoBERTa, 0.8792) +- (0, 0.0060)
          (DeBERTa, 0.9068) +- (0, 0.0041)
        };
        \addplot[black,fill=orange,error bars/y dir=both,error bars/y explicit] coordinates {
          (DistilBERT, 0.7244) +- (0, 0.0061)
          (BERT, 0.7484) +- (0, 0.0079)
          (RoBERTa, 0.7924) +- (0, 0.0147)
          (DeBERTa, 0.8292) +- (0, 0.0048)
        };
        \addplot[black,fill=violet,error bars/y dir=both,error bars/y explicit] coordinates {
          (DistilBERT, 0.6176) +- (0, 0.0092)
          (BERT, 0.6492) +- (0, 0.0093)
          (RoBERTa, 0.7200) +- (0, 0.0099)
          (DeBERTa, 0.7660) +- (0, 0.0081)
        };
        \addplot[black,fill=lime,error bars/y dir=both,error bars/y explicit] coordinates {
          (DistilBERT, 0.6216) +- (0, 0.0098)
          (BERT, 0.6832) +- (0, 0.0223)
          (RoBERTa, 0.6776) +- (0, 0.0075)
          (DeBERTa, 0.7208) +- (0, 0.0059)
        };
        \addplot[black,fill=pink,error bars/y dir=both,error bars/y explicit] coordinates {
          (DistilBERT, 0.5548) +- (0, 0.0056)
          (BERT, 0.6028) +- (0, 0.0076)
          (RoBERTa, 0.662) +- (0, 0.0058)
          (DeBERTa, 0.6888) +- (0, 0.0073)
        };
      \end{axis}
    \end{tikzpicture}
    \begin{tikzpicture}
      \begin{axis}[
        title={Tip Inference},
        xlabel={},
        ylabel={},
        set layers,
        ybar=1.0pt,
        bar width=4.5pt,
        symbolic x coords={DistilBERT, BERT, RoBERTa, DeBERTa},
        grid=both,
        ymin=0,
        legend cell align=left,
        enlarge x limits=0.15,
        ymin=0.5,
        ymax=1,
        x label style={font=\footnotesize},
        y label style={font=\footnotesize},
        ticklabel style={font=\footnotesize},
        legend image post style={scale=1.00},
        legend style={at={(1.05,0.20)},anchor=west,font=\scriptsize},
        ]
        \addplot[black,fill=red,error bars/y dir=both,error bars/y explicit] coordinates {
          (DistilBERT, 0.8748) +- (0, 0.0065)
          (BERT, 0.8980) +- (0, 0.0091)
          (RoBERTa, 0.9100) +- (0, 0.0034)
          (DeBERTa, 0.9368) +- (0, 0.0048)
        };
        \addlegendentry{Original}
        \addplot[black,fill=orange,error bars/y dir=both,error bars/y explicit] coordinates {
          (DistilBERT, 0.744) +- (0, 0.0085)
          (BERT, 0.7636) +- (0, 0.0095)
          (RoBERTa, 0.8084) +- (0, 0.0104)
          (DeBERTa, 0.8424) +- (0, 0.0115)
        };
        \addlegendentry{Synonym}
        \addplot[black,fill=violet,error bars/y dir=both,error bars/y explicit] coordinates {
          (DistilBERT, 0.6624) +- (0, 0.0154)
          (BERT, 0.69) +- (0, 0.0108)
          (RoBERTa, 0.7436) +- (0, 0.0046)
          (DeBERTa, 0.8072) +- (0, 0.0144)
        };
        \addlegendentry{Drop}
        \addplot[black,fill=lime,error bars/y dir=both,error bars/y explicit] coordinates {
          (DistilBERT, 0.6376) +- (0, 0.0110)
          (BERT, 0.6776) +- (0, 0.0075)
          (RoBERTa, 0.6448) +- (0, 0.0132)
          (DeBERTa, 0.7292) +- (0, 0.0073)
        };
        \addlegendentry{Placeholder}
        \addplot[black,fill=pink,error bars/y dir=both,error bars/y explicit] coordinates {
          (DistilBERT, 0.6228) +- (0, 0.0094)
          (BERT, 0.662) +- (0, 0.0058)
          (RoBERTa, 0.6632) +- (0, 0.0124)
          (DeBERTa, 0.7176) +- (0, 0.0059)
        };
        \addlegendentry{BERT Prediction}
      \end{axis}
    \end{tikzpicture}
}
\caption{Model performances tested on manipulated test data.}
\label{figure:ablation_study_on_keyword_manipulation}
\end{centering}
\end{figure*}
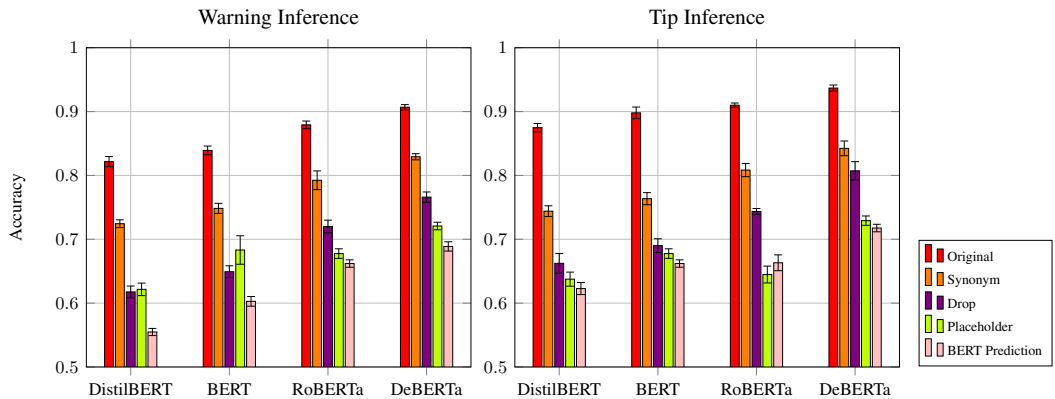

In addition to dropping, we manipulate the common keywords in the candidates using the following methods, as exemplified in Table \ref{table:keyword_manipulation_examples}:
    \begin{enumerate}
    \item Synonym Replacement: We replace such keywords with their synonyms using WordNet \citep{miller-1994-wordnet} with NLTK \citep{bird2009natural}.
    
    \item Replacing with a Placeholder: We replace such keywords with a placeholder word, which is simply \textsc{Placeholder}.
    
    \item Replacing with the BERT Prediction: We mask such keywords out with the [MASK] token and use BERT to predict the most likely token other than the original one.
    \end{enumerate}
    
    \begin{table*}
        \begin{centering}
        \scalebox{0.55}{
        \begin{tabular}{lcll}
            \textbf{Manipulation} & & \textbf{Positive Candidate (abridged)} & \textbf{Negative Candidate (abridged)} \\
            \toprule
            \textsc{Goal:} Stop Eating Fast Food & & &\\
            \textsc{Common Keywords:} Fast, Food, Eating & & &\\
            \midrule
            \multirow{2}*{Original} & & Beating a \textcolor{red}{fast food} addiction is a lot easier when you're not & Being vulnerable about your \textcolor{red}{eating} disorder is a hard thing to do. \\ 
            & & going it alone. Talk to friends about your goals for your diet. & Remember that you need other people to support you. \\ 
            \multirow{2}*{Synonym} & & Beating a \textcolor{orange}{flying food} addiction is a lot easier when you're not & Being vulnerable about your \textcolor{orange}{eat} disorder is a hard thing to do.\\ 
            & & going it alone. Talk to friends about your goals for your diet. & Remember that you need other people to support you.\\ 
            \multirow{2}*{Dropping} & & Beating a \textcolor{violet}{} addiction is a lot easier when you're not & Being vulnerable about your \textcolor{violet}{} disorder is a hard thing to do. \\ 
            & & going it alone. Talk to friends about your goals for your diet. & Remember that you need other people to support you. \\ 
            \multirow{3}*{Placeholder} & & Beating a \textcolor{green}{placeholder placeholder} addiction is a lot easier & Being vulnerable about your \textcolor{green}{placeholder} disorder is a hard \\ 
            & & when you're not going it alone. Talk to friends about your & thing to do. Remember that you need other people to \\ 
            & & goals for your diet. & support you. \\ 
            \multirow{2}*{BERT Prediction} & & Beating a \textcolor{pink}{new drug} addiction is a lot easier when you're not & Being vulnerable about your \textcolor{pink}{anxiety} disorder is a hard thing to do. \\ 
            & & going it alone. Talk to friends about your goals for your diet. & Remember that you need other people to support you. \\ 
            \bottomrule
        \end{tabular}}
        \captionsetup{justification=raggedright,singlelinecheck=false}
        \caption{Examples of keyword manipulation for a pair from the tip inference dataset. The goal is "Stop Eating Fast Food", which shares the common keywords of "Fast" and "Food" with the positive candidate and "Eating" with one of the negative candidates.}
        \label{table:keyword_manipulation_examples}
        \end{centering}
    \end{table*}
    
     We evaluate fine-tuned PLMs on these manipulated examples. As seen in Fig. \ref{figure:ablation_study_on_keyword_manipulation}, synonym replacement causes the least decrease in the performance with an approximate average of 10\% drop in accuracy across models and tasks. Dropping, placeholder replacement, and BERT prediction replacement closely follow each other respectively, with average decreases in accuracy varying from 15\% to 20\%. 
     
\label{sec:additional_keyword_manipulation_methods}

\section{DeBERTa and GPT-4 Failures}
Specific examples that DeBERTa and GPT-4 fail can be seen in Table \ref{table:deberta_gpt4_fails}.

\begin{table*}
    \begin{centering}
    \scalebox{1.00}{
    \begin{tabular}{lcr}
        \textbf{DeBERTa} & & \textbf{GPT-4} \\
        \toprule
        Install Ceramic Wall Tile &  & Block People on Facebook \\
        Build a Nerf Fort &  & See Your WiFi Password on an iPhone \\
        Turn a Cardboard Box Into a Basket &  & Change the Password in Outlook 365 \\
        Prevent Water Stains on Bathroom Walls &  & Deal with Catty Coworkers \\
        Make a Rope Braid &  & Be Confident As a Short Person \\
        Remove a Red Wine Stain Ring from a Wood Table && Help an Autistic Family Member \\
        \bottomrule
    \end{tabular}}
    \caption{Goals of examples that DeBERTa and GPT-4 fail.}
    \label{table:deberta_gpt4_fails}
    \end{centering}
\end{table*}
\label{sec:deberta_gpt4_fails}

\section{A Comparison Between Warnings and Tips}
As discussed in Sec. \ref{subsection:cross_tests}, warnings and tips share the common purpose of presenting additional information to user for better instruction execution within procedural wikiHow tutorials. Thus, they generally have high similarity in semantics and structure. 

Overall, warning inference poses a greater challenge compared to tip inference. We believe this is due to two main reasons. First, tips are more goal-specific, while warnings are more general, yet still distinguishable. For example, any goal that contains sharp objects might have a warning towards using those sharp objects carefully; yet, the same does not hold true for tips. Second, tips are ample in quantity; therefore, there are more examples to learn from.

Our reasoning behind tips being more goal-specific and informative regarding the procedural tutorial compared to warnings is reinforced by the following findings:

\begin{enumerate}

\item Both PLMs and LLMs perform better in tip inference compared to warning inference, indicating a greater capability in reasoning with tips due to tips being more directly connected to their goals.
\item Previous fine-tuning on tip inference better improves the performance in transfer learning compared to previous fine-tuning on warning inference, demonstrating that tip inference contributes to model's learning over procedural tasks more.
\item Tip inference is affected more by keyword manipulation, showing that tip inference data is more vulnerable to keyword alteration because its positive candidates contain more goal-specific vocabulary.

\end{enumerate}
\label{sec:warnings_vs_tips}

\end{document}